\documentclass[letterpaper, 10 pt, conference, final]{ieeeconf}  % Comment this line out if you need a4paper

\IEEEoverridecommandlockouts                              % This command is only needed if 
\usepackage{array,multirow}
\usepackage{hyperref}
\usepackage{tabularx}
\usepackage{graphicx}  
\usepackage{subcaption} 
\usepackage{tcolorbox}

\usepackage{bmpsize}
\usepackage{amsmath}
\usepackage{pifont}
\usepackage{amssymb}

\usepackage[commandnameprefix=always]{changes}
\definechangesauthor[name={Karla}, color=teal]{KS}
\definechangesauthor[name={Rado}, color=orange]{RS}
\definechangesauthor[name={Petr}, color=brown]{PV}

\title{\LARGE \bf
TransforMerger: Transformer-based Voice-Gesture Fusion for Robust Human-Robot Communication}
\author{
Petr Vanc$^{1}$
\and 
Karla Stepanova$^{1}$
\thanks{$^{1}$Czech Technical University in Prague, Czech Institute of Informatics, Robotics, and Cybernetics, \texttt{petr.vanc@cvut.cz}, \texttt{karla.stepanova@cvut.cz}}
\thanks{  This work was supported by the European Union under the project Robotics and advanced industrial production (reg. no. CZ.02.01.01/00/22\_008/0004590), and by the Czech Science Foundation (project no. GA21-31000S). P.V. by CTU Student Grant Agency (reg. no. SGS23/138/OHK3-027/23).}}

\begin{document}

\maketitle
\thispagestyle{empty}
\pagestyle{empty}

%%%%%%%%%%%%%%%%%%%%%%%%%%%%%%%%%%%%%%%%%%%%%%%%%%%%%%%%%%%%%%%%%%%%%%%%%%%%%%%%
\begin{abstract}
As human-robot collaboration advances, natural and flexible communication methods are essential for effective robot control. Traditional methods relying on a single modality or rigid rules struggle with noisy or misaligned data as well as with object descriptions that do not perfectly fit the predefined object names (e.g. 'Pick that red object'). 
We introduce TransforMerger, a transformer-based reasoning model that infers a structured action command for robotic manipulation based on fused voice and gesture inputs. Our approach merges multimodal data into a single unified sentence, which is then processed by the language model. We employ probabilistic embeddings to handle uncertainty and we integrate contextual scene understanding to resolve ambiguous references (e.g., gestures pointing to multiple objects or vague verbal cues like "this").
We evaluate TransforMerger in simulated and real-world experiments, demonstrating its robustness to noise, misalignment, and missing information. Our results show that TransforMerger outperforms deterministic baselines, especially in scenarios requiring more contextual knowledge, enabling more robust and flexible human-robot communication.
Code and datasets are available at: \url{http://imitrob.ciirc.cvut.cz/publications/transformerger}.
%
%Traditional approaches often rely on a single modality or predefined rigid rules, making them susceptible to errors from misaligned or noisy data. In this paper, we introduce TransforMerger, a novel transformer-based reasoning model that fuses multimodal inputs—voice and gestures—into a structured Skill Command for robotic manipulation. At its core, we leverage a state-of-the-art transformer-based model as the reasoning engine, enabling it to process probabilistic soft embeddings from token distributions and effectively handle uncertain or conflicting inputs. The model integrates contextual scene understanding and object properties to disambiguate vague references, such as gestures pointing to multiple objects or ambiguous verbal cues like "this" or "that." We evaluate our method on both simulated and real-world experiments, demonstrating its robustness to noisy data, misalignment, and missing information.  Our results show that TransforMerger outperforms deterministic baselines and zero-shot LLM approaches, achieving higher accuracy in intent recognition and multimodal alignment. Code and datasets are publicly available on our project website: http://imitrob.ciirc.cvut.cz/publications/transformerger.

 \end{abstract}

\maketitle

\section{Introduction}
Human communication integrates multiple modalities—language, gestures, gaze, and facial expressions—ensuring robustness against missing, noisy, or conflicting information. Context and background knowledge further enhance understanding, enabling efficient interaction.

In contrast, human-robot interaction (HRI) often relies on rigid communication constrained to single modalities (e.g., language \cite{pires2005robot}, gestures \cite{Vanc2023}) or strictly partitioning modalities role (e.g., language defines actions, gestures specify locations). Existing multimodal approaches often naively fuse inputs, limiting their adaptability~\cite{Wang_2024}.

\begin{figure}
    \centering
    \includegraphics[trim={0.75cm 0.75cm 0.75cm 0.75cm},clip,width=1.0\linewidth]{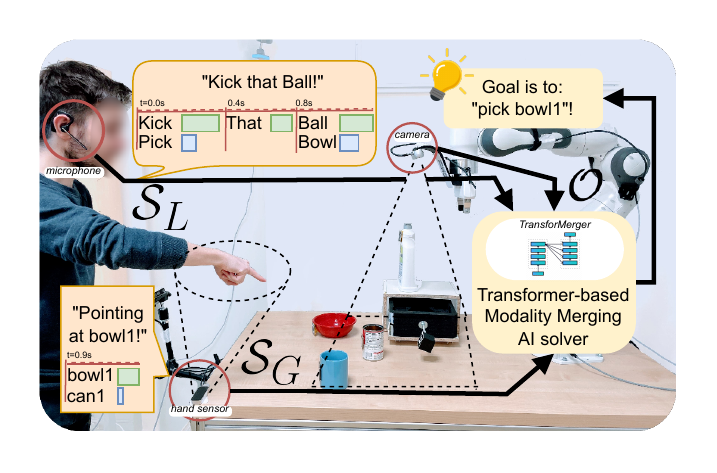}
    \caption{Human-Robot Interaction Pipeline. A user communicates tasks through hand gestures (\(\mathcal{S}_G\)), captured via a hand sensor, and voice commands (\(\mathcal{S}_V\)), recorded by a microphone. A camera monitors the scene to get scene objects (\(\mathcal{O}\)). Our solution utilizes a transformer-based SOTA Large Language model (1-3B param., running offline) to reason about the user's intent and generate clear action commands for the robot to execute.}
    \label{fig:intro}
\end{figure}

To bridge this gap, we propose a context-aware multimodal merging algorithm incorporating transformer-based large language models \cite{huggingfacetransformers}. Our approach dynamically integrates uncertain multimodal inputs, updating action probabilities based on simultaneous observations (see Fig.~\ref{fig:intro}). This allows the system to resolve ambiguity, assess action feasibility, and improve robustness to noise and misalignment.
We evaluate our method on simulated and real-world datasets, testing alignment, noise levels, and action complexity. Some datasets contain conflicting multimodal information, requiring contextual resolution. %Ablation studies assess the contribution of key system components.

%Human communication integrates multiple modalities—vision, language, gestures, eye gaze, and facial expressions—enabling robustness against missing, noisy, or conflicting information. Context and background knowledge further enhance understanding, ensuring efficient interaction.

%In contrast, current human-robot interaction (HRI) setups often impose rigid communication, either relying on a single modality (e.g., language \cite{pires2005robot}, gestures \cite{Vanc2023}) or strictly partitioning modalities into specific roles. For example, language typically defines actions, while gestures or gaze specify parameters like location or movement direction. Existing multimodal approaches often integrate information naively, limiting their adaptability.

%A more general approach is needed to effectively merge multimodal data and accurately infer human intent. This requires not only fusing uncertain information but also assessing the feasibility of actions—such as whether the robot can reach, grasp, or interact with an object appropriately. Sensor fusion techniques provide a foundation, but they must be adapted for HRI to ensure natural and reliable collaboration.

%In this paper, we take inspiration from the recent large language model development that are powered by transformers \cite{huggingfacetransformers} and tuned for instruction following and reasoning~\ref{TODO} and propose a merging algorithm that provides a robust, context-aware solution for combining information from various modalities (see Fig.~\ref{fig:intro}). Our approach handles multiple beliefs over possible actions from different modalities, updating the probability of these actions and their parameters by simultaneously combining information. 

%We evaluate our proposed approach using well-controlled artificial data from gesture and language modalities, incorporating varying levels of alignment, noise, and action complexity. Some datasets contain data with conflicting information in both modalities, only resolvable based on context. Multiple ablation studies highlight the importance of different parts of the system.

In summary, the main contributions of the paper are:
\begin{itemize}
    \item TransforMerger (see Fig.~\ref{fig:system}), a context-aware model for merging multimodal data, showing improved robustness to noise, misalignment and capable of resolving input ambiguities using contextual knowledge and by grounding object attributes in scene context (e.g., identifying 'red metal object').
\item An evaluation on simulated and real-world dual-modality (gesture and language) datasets, analyzing the impact of different noise types. We compare three state-of-the-art language models as reasoning engines against a deterministic baseline.
\end{itemize}

Datasets, code, and models are on the project website\footnote{\label{projectwebsiteref}Project website:
\href{http://imitrob.ciirc.cvut.cz/publications/transformerger}{http://imitrob.ciirc.cvut.cz/publications/transformerger}}.

\section{Related work}

Recent research has explored multimodal fusion, integrating speech, gestures, and gaze \cite{Wang_2024}, but these approaches often treat modalities separately rather than as a unified representation, limiting their ability to resolve ambiguity.

A key challenge in multimodal perception is handling uncertainty arising from sensor noise, speech recognition errors, and ambiguous gestures. Traditional probabilistic models, such as Bayesian networks \cite{bishop2006pattern}, hidden Markov models (HMMs) \cite{rabiner1989tutorial}, and probabilistic graphical models \cite{starner1998visual}, have been employed to mitigate these issues. However, these methods rely on predefined rules and do not incorporate contextual reasoning, making them ineffective in cases where multiple references (e.g., pointing gestures) lack explicit grounding.

Recent advances in large language models (LLMs) have introduced powerful reasoning capabilities for context-aware decision-making \cite{gpttrasnformersbrown2020}. Zero-shot and few-shot learning techniques \cite{wei2022finetunedlanguagemodelszeroshot} allow LLMs to generalize beyond fixed rule-based systems, enabling them to infer missing or ambiguous information from broader context. However, most LLM-based HRI systems remain limited to text-based interactions, failing to fully integrate gesture-based communication into their reasoning processes. This highlights the need for a more comprehensive approach that merges both spoken and non-verbal cues in a probabilistic manner.
Transformer-based architectures have demonstrated state-of-the-art performance in multimodal learning, particularly in vision-language \cite{dosovitskiy2021imageworth16x16words} and speech recognition \cite{radford2022robustspeechrecognitionlargescale}. However, existing models like CLIP and Flamingo \cite{NEURIPS2022_960a172b} primarily focus on aligning image and text data, lacking the ability to merge gesture-based inputs with speech for robotic manipulation tasks. This gap motivates the development of a system that leverages transformer-based reasoning to align multimodal inputs while accounting for temporal misalignment and uncertainty.

Recent works have advanced multimodal HRI by integrating gesture, speech, and contextual reasoning for improved interaction. Wang et al. \cite{wang2024thisthatlanguagegesturecontrolledvideo} explored language-gesture conditioned video generation for robotic planning, while Trick et al. \cite{trick2019multimodaluncertaintyreductionintention} proposed probabilistic fusion of gaze, gestures, and speech to reduce ambiguity in intention recognition. Ferrari et al. \cite{ferrari2024collaborativeconversationsafemultimodal} addressed safety in HRI by fusing gesture and speech through tensor-based concatenation for risk-aware collaboration. The closest work to ours is \cite{lai2025nmmhrinaturalmultimodalhumanrobot}, which uses GPT-4 to process language input and determine target action and its parameters. This information is fused by LLM with objects selected by deictic gestures to determine the robotic action and its parameters.

\textbf{TransforMerger} extends these methods by incorporating a state-of-the-art transformer model for probabilistic reasoning. Unlike previous approaches, our model explicitly integrates probabilistic inputs from both modalities along with context-specific parameters such as scene descriptions. This allows the model not only to enhance contextual recognition of individual objects but also to account for noise in the inputs as well as for the temporal misalignment between deictic gestures and voice commands. As a result, TransforMerger enables more natural and robust robotic action generation.
%TO ADD: 2024 This and that - Language-gesture conditioned video generation -\cite{wang2024thisthatlanguagegesturecontrolledvideo} 
%- 2025 - NMM-HRI: Natural Multi-modal Human-Robot Interaction with Voice and Deictic Posture via Large Language Model \cite{lai2025nmmhrinaturalmultimodalhumanrobot}
%- Multimodal Uncertainty Reduction for Intention Recognition in Human-Robot Interaction [\cite{trick2019multimodaluncertaintyreductionintention}] -  fusion of gaze + gestures + speech + objects
%- possibly: Collaborative Conversation in Safe Multimodal Human-Robot Collaboration
%- gesture + voice -> fusion -> safety
%- fusion as a concatenation of vectors inside a tensor
%- concerned about safety layer
%- \cite{ferrari2024collaborativeconversationsafemultimodal}

\section{Problem formulation}
\label{sec:problem_formulation}

The goal of this work is to infer a structured Skill Command that encapsulates a robotic manipulation action and its parameters (Sec.~\ref{sec:skill_command}). This high-level instruction is derived from multimodal inputs, primarily gestures and voice. We leverage large language models (LLMs) to integrate knowledge from multiple modalities (Sec.~\ref{sec:modality_preprocessing}) and generate an executable Skill Command.

Unlike traditional methods that rely on predefined commands or deterministic parsing, our approach incorporates probabilistic reasoning to handle the ambiguities and uncertainties inherent in human communication. Since gesture and speech inputs are often noisy, incomplete, or ambiguous, we represent them in a probabilistic format, allowing the system to reason over multiple interpretations and select the most probable command.

Each input modality undergoes independent preprocessing to achieve a uniform representation (Sec.~\ref{sec:modality_preprocessing}). Since modalities are inherently noisy, we do not assume perfect recognition but instead propagate uncertainty probabilistically, enabling the system to make robust decisions.

This chapter formalizes the problem by first stating the Main Objective (Sec.~\ref{sec:main_objective}), then describing probabilistic modality representation and multimodal fusion (Sec.~\ref{sec:probabilistic_modality_representation}), followed by defining the Skill Command syntax and execution model (Sec.~\ref{sec:skill_command}).

\subsection{Main Objective}
\label{sec:main_objective}

%The objective is to merge different modalities, focusing specifically on hand gestures and natural language. While our approach is demonstrated using these two modalities, it could also be extended to incorporate eye gaze for pointing or body language. The merging process involves selecting the most probable interpretation from both modalities and mapping it to a structured Skill Command for execution. Given probabilistic representations of gestures ($\mathcal{S}_G$) and voice ($\mathcal{S}_V$)—both of variable length—the system should be able to: First, infer the most probable meaning from multimodal inputs. Second, resolve ambiguities by leveraging confidence scores from each modality. Third, select an executable Skill Command based on the detected intent. Finally, consider the set of available objects $\mathcal{O}$, ensuring commands reference physically present entities. In the next section, we describe the gesture and language preprocessing pipeline.
The objective is to merge multimodal inputs—hand gestures and natural language—to infer a structured Skill Command for execution. While demonstrated with two modalities, the approach could extend to other modalities such as eye gaze or body language. Given probabilistic representations of gestures ($\mathcal{S}_G$) and voice ($\mathcal{S}_V$), the system must: First, infer the most probable meaning from noisy and ambiguous multimodal inputs. Second, resolve ambiguities using confidence scores from each modality. Third, select an executable Skill Command based on the detected intent. Finally, ensure that referenced objects ($\mathcal{O}$) are physically present and correctly identified. In the Section~\ref{sec:modality_preprocessing}, we describe the gesture and language preprocessing pipeline.

\subsection{Probabilistic Modality Representation}  
\label{sec:probabilistic_modality_representation}

We define a unified probabilistic representation for each modality $m$, treating each input sentence as a sequence of words:

\begin{equation}
\mathcal{S}_{m} = (\mathbf{w}_1, \ldots, \mathbf{w}_{n})
\end{equation}

In our case, we consider two modalities, gestures ($G$) and language ($V$):
\begin{equation}
\mathcal{S}_G = (\mathbf{w}_1, \ldots, \mathbf{w}_k), \quad \mathcal{S}_V = (\mathbf{w}_1, \ldots, \mathbf{w}_l)   
\end{equation}

Each word $\mathbf{w}$ is associated with a timestamp $t$ and a probability distribution over possible interpretations:
\begin{equation}    
\mathbf{w} = \{(t, w_i, \mathcal{P}_i) \mid i = 1, \dots, N_k\}
\label{eq:word}
\end{equation}

where $w_i$ is a word candidate and $\mathcal{P}_i$ its probability, $N_k$ is number of candidates for the $k$-th word. For example, in speech recognition the model may output:
\begin{equation}
\mathbf{w} = (t=0.1, \{\text{"pick"}: 0.9, \text{"kick"}: 0.1\})
\end{equation}

indicates "pick" is the most probable interpretation but with some uncertainty. A similar probabilistic mapping is applied in gesture recognition, linking gestures to discrete action words with associated probabilities.
\subsection{Skill Command}
\label{sec:skill_command}

We define a \textit{Skill Command} in a deterministic format that encapsulates user intent, enforcing the reasoning model to transform multimodal input into an interpretable command for robotic execution.
%. This format serves as an intermediate representation, enforces the reasoning model to translate the multimodal input into a structured command that the robot can interpret and execute.
%While this format does not capture all task variations (e.g., tool usage: "pick an object using the hammer"), it remains extensible to include additional parameters. 
A Skill Command follows a structured syntax passed to the LLM for reasoning. It must contain at least one required parameter (action), with additional parameters depending on the action type (e.g., "pour" requires a target object). In our evaluation, we use a syntax covering manipulation actions involving 0, 1, or 2 objects (Sec.~\ref{sec:actions_objects}). The syntax is easily extensible for domain-specific tasks (e.g., "Move bowl towards box using a hammer" would require an additional object parameter).

%A Skill Command has a structured syntax. The specific syntax is then passed to the LLM model within prompt for reasoning. For robotic manipulation tasks Skill Command should have always at least one required parameter (action). Depending on the type of actions, other parameters might be required (e.g., target object for action pour). In our evaluation we consider the following syntax that covers vast majority of manipulation actions with 0, 1 or two 2 involved objects (see \ref{sec:actions_objects} for examples of these actions used in our experiments). If different type of manipulation actions are expected in the given usecase, the Skill Command syntax can be easily adjusted (e.g. Move bowl towards box by hammer) would require extending the syntax by adding another object.

\begin{equation}  
\text{Skill Command} = \underbrace{ap}_{\texttt{mod.}} \oplus \underbrace{a}_{\texttt{action}} \oplus \underbrace{to1}_{\texttt{object}} \oplus \underbrace{p}_{\texttt{prep.}} \oplus \underbrace{to2}_{\texttt{object}}  
\end{equation}  

where:  
\begin{enumerate}
\item $ap$ (\textbf{action parameter}) adjusts execution parameters, e.g., speed or force (\textit{optional}).  
\item $a$ (\textbf{target action}) specifies the core robotic skill to be executed (\textit{required}).
\item $to1$ (\textbf{target object 1}) represents the primary object of interaction.  
\item $p$ (\textbf{preposition}) defines spatial or relational constraints between objects.  
\item $to2$ (\textbf{target object 2}) is the secondary object.
\end{enumerate}

$to1$ is required for some actions (with 1 or 2 involved objects), $p$ and $to2$ are required only if interacting with two objects. For example, a simple command instantiation could be:  

\begin{equation}  
\text{Skill Command} = \text{``quickly push tomatoes1 near bowl1''}  
\end{equation}  

Here, "quickly" modifies execution speed ($ap$), "push" specifies the action ($a$), "tomatoes1" is the target object ($to1$), "near" defines spatial context ($p$), and "bowl1" is the secondary object ($to2$).

\subsubsection{Command Interpretability \& Execution}  

Each Skill Command maps to predefined robotic skills, ensuring structured execution. The preposition $p$ and action parameter $ap$ directly modify trajectory execution, influencing speed, force, or object placement.  
The role of $to1$ and $to2$ is to specify scene objects, which are dynamically identified in the robot's environment. For instance, the command \textit{"put into"} results in a different execution trajectory than \textit{"put on top of"}, even though both involve placing an object.  
We learn these parameterized trajectories from demonstration (Sec.~\ref{sec:kinesthetic_demos}).
%Skill is a parametrized trajectory that we learn from demonstration, read the details in Sec.~\ref{sec:kinesthetic_demos}.

\section{Proposed Solution: TransforMerger}
\label{sec:proposed_solution}

\begin{figure}
    \centering
    \includegraphics[width = \linewidth]{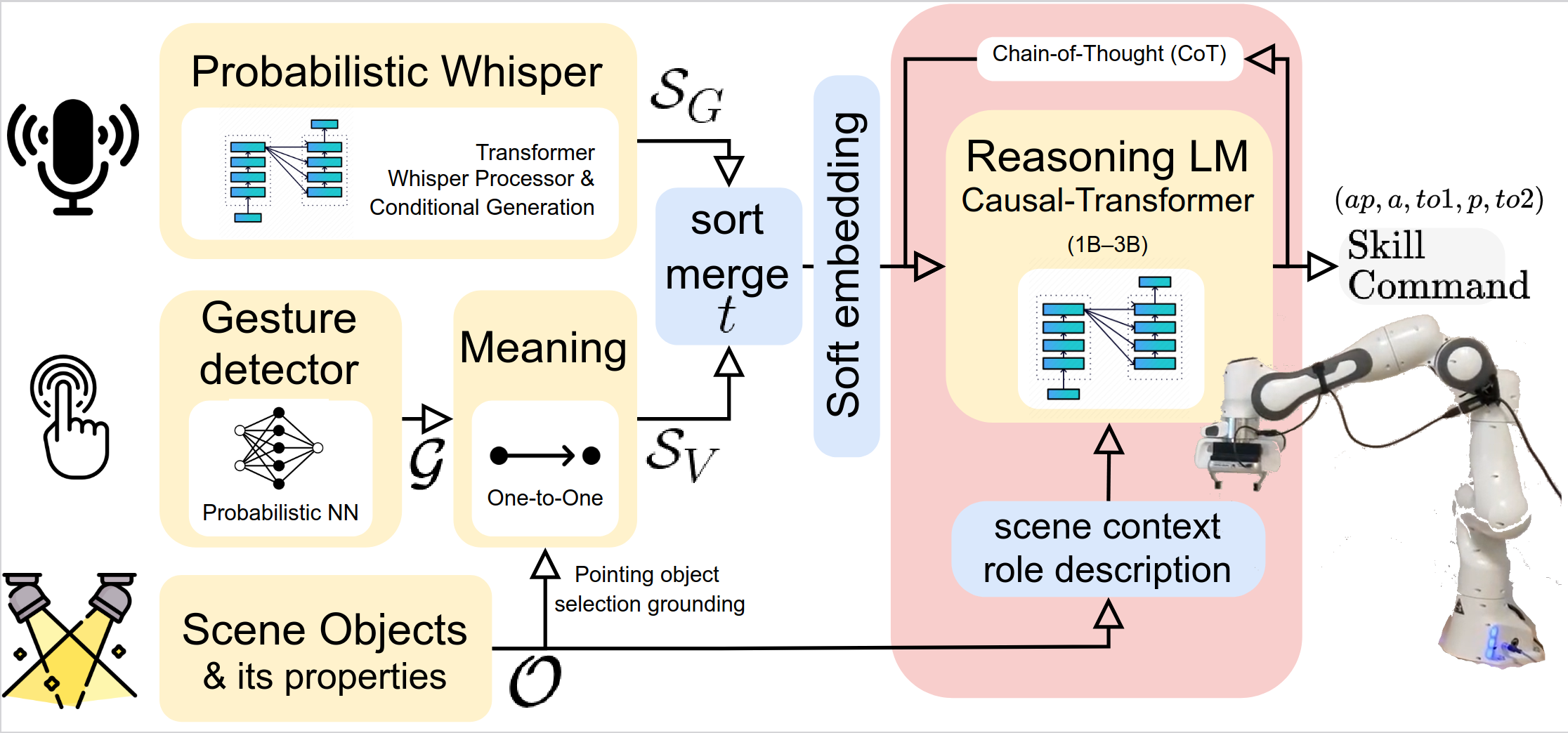}
    \caption{System architecture for real world experiments semantic reasoner from Fig.~\ref{fig:intro}. System is merging multimodal inputs into a single Skill Command, a high-level instruction for a robot to execute. The blue blocks highlight the paper contributions. %are paper contributions: The sort-merge, 2) soft embeddings for LM, 3) provided context to language model within the prompt. 
    In the simulated setup the $\mathcal{S}_G$ and $\mathcal{S}_V$ are simulated by created dataset, see Sec.~\ref{sec:artificial_dataset}. % $\mathcal{G}$ are detected gestures.
    }
    \label{fig:system}
\end{figure}

In this section, we introduce \textit{TransforMerger}, a novel approach for generating structured robotic Skill Commands from multimodal inputs—voice and gestures (see Fig.~\ref{fig:system}).

Our method processes multimodal inputs in a post-merging manner \cite{baltruvsaitis2018multimodal}, where preprocessed data from each modality is integrated into a unified probabilistic representation. This approach ensures a coherent fusion of gesture-based and voice-based instructions while maintaining temporal and contextual dependencies. 

Since multimodal inputs often exhibit misalignment and noise, preprocessing errors are inevitable. A core contribution of this work is demonstrating how TransforMerger mitigates these errors, improving the robustness of multimodal understanding and enhancing downstream task performance. The preprocessing steps applied in real-world experiments are detailed in Sec.~\ref{sec:modality_preprocessing}.

%The preprocessing steps applied in real-world experiments are detailed in Sec.~\ref{sec:modality_preprocessing}.

%Given the inherent noise and misalignment in multimodal inputs, preprocessing errors are inevitable. The core contribution of this work is demonstrating how our proposed approach effectively mitigates these errors, enhancing the robustness of multimodal understanding and improving downstream task performance.

\subsection{Merging algorithm}
\label{sec:merging_algorithm} 

\subsubsection{Merging Process}  
To integrate multimodal inputs, we concatenate sentences from individual modalities and sort them by timestamp to ensure a consistent temporal structure:

%The merging function concatenates the sentences from individual modalities and sorts them by their timestamps, ensuring a coherent temporal structure:

\begin{equation}
\mathcal{S}_M = \underset{\mathbf{t}}{\text{sort}}(\mathcal{S}_G \oplus \mathcal{S}_V),
\end{equation}

where $\oplus$ represents concatenation, and $\text{sort}(\cdot)$ arranges words based on their timestamps $\mathbf{t}$. Although this formulation focuses on merging gesture and voice inputs, it can be naturally extended to additional modalities. The structure of individual sentences ($\mathcal{S}_{{M,G,V}}$) is detailed in Sec.~\ref{sec:probabilistic_modality_representation}, in which each detected human word (gesture or voice) is represented as a weighted distribution over possible alternatives, see Eq.~\ref{eq:word}. This uniform sentence format allows us to discard modality-specific information and treat all inputs consistently (see Fig.~\ref{fig:merged} for an example of the merged inputs).
% ($\mathcal{S}_{\{M,G,V\}}$) in Sec.~\ref{sec:probabilistic_modality_representation}. This approach is possible as sentences in individual modalities share the same format. %, allowing us to discard modality-specific information and treat all inputs uniformly. 

\begin{figure}[!ht]
    \centering
    \begin{tcolorbox}[
        colback=gray!10, colframe=black, width=\linewidth, boxrule=0.8pt,
        left=2pt, right=2pt, top=2pt, bottom=2pt, colbacktitle=black!20
    ]
        \scriptsize
        
        {\bf Voice ($S_V$):} \texttt{[(0.3, {'place': 0.8, ‘plate’:0.3,...}), (0.5, {'cup': 0.6, ‘cap’:0.4,...}), (0.6, {'to': 1.0}), (0.9, {'cube': 0.5,’tube’:0.3,...})]
} \\
         {\bf Gestures ($S_G$):} \texttt{[(0.8, {'cup': 0.85, 'cube': 0.31, 'plate': 0.24, 'table': 0.01, 'can': 0.01, 'box': 0.01,...})]}\\
         {\bf Merged inputs ($S_M$):} \texttt{[(0.3, {'place': 0.8, ‘plate’:0.3,...}), (0.5, {'cup': 0.6, ‘cap’:0.4}), (0.6, {'to': 1.0}),(0.8, {'cup': 0.85, 'cube': 0.31, 'plate': 0.24, 'table': 0.01, 'can': 0.01, 'box': 0.01,...}) (0.9, {'cube': 0.5,’tube’:0.3})]
}    
    \end{tcolorbox}
    \caption{Example simulated inputs from gesture and language and the result of their merging (see Sec.~\ref{sec:merging_algorithm}).}
    \label{fig:merged}
\end{figure}

The merging process introduces several challenges:
1) \textit{Duplicate words} may appear when gestures and speech reference the same content simultaneously. The reasoning language model (Sec.~\ref{sec:causal_transformer}) must detect and filter these redundancies to infer the correct interpretation.
2) \textit{Voice-gesture asynchrony} makes direct alignment between modalities difficult.
3) \textit{Ambiguous references} voice commands such as "this", "the red object", "that", etc., lack explicit grounding, making it unclear which object is being referred to. Additionally, if a user says "this" while pointing between two objects, it remains uncertain whether they refer to one or both.

These challenges make merging input relations with corresponding modality representations non-trivial and necessitate \textit{context-aware reasoning} to resolve ambiguities.
To address these issues, our approach incorporates reasoning language models—the core component of our system (Sec.~\ref{sec:causal_transformer})—to infer contextual grounding and resolve ambiguities in multimodal inputs. While not all misalignment cases are explicitly analyzed, our experiments (Sec.~\ref{sec:experiment_noise_align}) demonstrate that \textit{TransforMerger} effectively mitigates common errors in noisy, ambiguous, and asynchronous multimodal scenarios.

%However, this can result in duplicate words when gestures and speech reference the same content simultaneously. The reasoning language model (Sec.~\ref{sec:causal_transformer}) must account for these redundancies to infer the correct interpretation.

%Due to the inherent asynchrony between voice and gestures, direct alignment between them is often infeasible. Gesture-based references, such as "this" or "that," frequently lack explicit grounding, making it ambiguous which object they refer to. Additionally, discrepancies in the number of references across modalities further complicate interpretation. For instance, if a user says "this" while pointing at two different objects, it remains unclear whether the reference pertains to the first, the second, or both. Consequently, merging input relations with the corresponding modality representations is not straightforward. 

%To address this, we demonstrate that a language model (LM), the core part of our system, introduced in Sec.~\ref{sec:causal_transformer}, effectively reasons about contextual grounding to resolve such ambiguities (see experiments in Sec.\ref{sec:experiment_noise_align}).

\subsection{Foundational reasoning model with soft embeddings}  
\label{sec:causal_transformer}  

We employ state-of-the-art instruct-tuned language models with reasoning capabilities, built on a Causal Transformer architecture \cite{gpttrasnformersbrown2020} (see Sec.~\ref{sec:model_comparison} for the specific models used).  The language model processes merged probabilistic inputs (see Sec.~\ref{sec:merging_algorithm} and Fig.~\ref{fig:merged}) and infers the most likely object reference based on the context. We introduce soft embeddings to enforce the model to reason probabilistically over the inputs and design a parametrized prompt to constrain the model and provide contextual information.

\begin{figure}[t]
    \centering
    \begin{tcolorbox}[
        colback=gray!10, colframe=black, width=\linewidth, boxrule=0.8pt,
        left=2pt, right=2pt, top=2pt, bottom=2pt, colbacktitle=black!30
    ]
        \scriptsize
        
        \tcbsubtitle{Task}         \scriptsize
        You are an \textbf{assistant} that analyzes user requests to infer \textit{actions}, \textit{objects}, \textit{relationships}, and \textit{action property}. Follow these steps:

        \tcbsubtitle{Reasoning Steps}
        \begin{enumerate}
            \item Read the user’s input.
            \item Identify the action (from: \texttt{\scriptsize<inserted\_actions>}) and its property (e.g., speed: "fast"). If actions/property are repeated (e.g., 'fast fast pour'), treat them as stronger evidence for a single instance (e.g., 'fast').
            \item Determine the primary object (from: \texttt{\scriptsize<inserted\_objects>}). If objects are mentioned multiple times (e.g., 'cup cup'), infer they refer to the same grounded instance (e.g., \texttt{cup1}), unless attributes/context imply separate objects.
            \item Check for a secondary object and its relationship to the primary object (e.g., "to", "from").
            \item Explain reasoning, check the valid actions and objects. Verify if repeated terms map to a single object instance in the scene. If ambiguity exists, use attributes or default to the primary valid object.
            \item Output your reasoning, then finalize with \textbf{output} in the following format:
            \begin{center}
                \fbox{\parbox{0.9\linewidth}{
                \scriptsize\texttt{action: X, object1: Y, object2: Z, property: P, relationship: R}
                }}
            \end{center}
        \end{enumerate}

        \tcbsubtitle{Context}
        \begin{itemize}
            \item {\bf Valid properties:} \texttt{\scriptsize<inserted\_properties>}
            \item {\bf Valid Actions:} \texttt{\scriptsize<inserted\_actions>}
            \item {\bf Valid Objects:} \texttt{\scriptsize<inserted\_objects>}
            \item {\bf Scene Description:} \texttt{\scriptsize<inserted\_scene\_description>}
        \end{itemize}

    \tcbsubtitle{Examples}
Example 1: Simple Action \\
\textbf{User}: "pick up cup1 cup."  \\
\textbf{Assistant}:  
`action: pick, object1: cup1, object2: none, property: none, relationship: none`\\

Example 2: Action with Property\\
\textbf{User}: "slow pour cup cup1 to bowl1 bowl."  \\
\textbf{Assistant}:  
`action: pour, object1: cup1, object2: bowl1, property: slow, relationship: to`\\

Example 3: Attribute-Based Object\\
\textbf{User}: "pick up the wide blue object."  \\
\textbf{Assistant}:  
`action: pick, object1: cube1, object2: none, property: none, relationship: none`

    \end{tcolorbox}
    \caption{System prompt: Parameterized scene-aware prompt for the reasoning model, incorporating structured reasoning steps, model's role and required output. Available actions, objects, and scene descriptions are dynamically inserted as parameters for each specific task (example in Fig.~\ref{fig:example_paraemters}).}
    \label{fig:reasoning_prompt}
\end{figure}

\begin{figure}[t]
    \centering
    \begin{tcolorbox}[
        colback=gray!10, colframe=black, width=\linewidth, boxrule=0.8pt,
        left=2pt, right=2pt, top=2pt, bottom=2pt, colbacktitle=black!20
    ]
        \scriptsize
        
        {\bf Valid Properties:} \texttt{[fast, slow, carefully, force]} \\
         {\bf Valid Actions:} \texttt{["stop", "release", "pick", "push", "pass",
"place", "open", "close", "pour"]}\\
         {\bf Valid Objects:} \texttt{["cup", "cube", "plate", "table", "box"]}\\
         {\bf Scene Description:} \texttt{cube is a small red cube. cup is a medium red cup. plate is a small blue plate. box is a big black box.}
    
    \end{tcolorbox}
    \caption{Example task parameters (objects, actions, and scene description) inserted to the system prompt (Fig.~\ref{fig:reasoning_prompt}) for one of the experiments.}
    \label{fig:example_paraemters}
    \vspace{-0.4cm}
\end{figure}

% \begin{figure}
%    \centering
%    \includegraphics[width=1.0\linewidth]{images/roledescription.drawio.pdf}
%    \vspace{-2.5em}
%    \caption{Parameterized scene-aware prompt for the reasoning model, incorporating structured reasoning steps. It defines the model's role, required output, and available actions, objects, and scene details. Actions, objects, and scene descriptions are dynamically inserted as parameters for each specific task.
%     Role description for the Reasoning model, where actions, objects, and scene description is inserted.
%    }
%    \label{fig:role_description}
% \end{figure}

\subsubsection{System prompt with parametrized structured reasoning for optimized performance}  
\label{sec:reasoning_strategies}  
The system prompt (see Fig.~\ref{fig:reasoning_prompt}) consists of the task specification, the model’s role, structured reasoning steps, and parameterized contextual information. The contextual parameters include available actions ($A$), objects ($O$), properties ($P$), and the scene description ($S$), which collectively constrain the model to the specific task. 
  %  $$
  %  \text{Role description} = (A, O, P, S)
  %  $$

To enhance reasoning strategies, we refine the model's role description to guide its inherent reasoning capabilities. Rather than directly predicting the final Skill Command, we leverage the model’s common-sense reasoning to infer context-aware decisions. Additionally, the model is instructed to expect class objects but return their real-world instances. Finally, we enforce a structured output format, ensuring adherence to the predefined Skill Command syntax.
For the exact prompt used, see the project website\textsuperscript{\ref{projectwebsiteref}}.
%    System prompt design is composed of a role description (Fig.~\ref{fig:role_description}), which defines the intended function, responsibilities, and constraints for our application. It is fed by available actions $A$, objects $O$, properties $P$, and Scene classes $S$:

\subsubsection{Soft embeddings} Our contribution to the reasoning model is the construction of soft embeddings to enforce probabilistic reasoning. We begin by processing probabilistic inputs, tokenizing them into text units, and computing weighted representation that captures transcription uncertainty. For each token candidate, the system tokenizes the string into subword tokens, retrieves their embeddings, and computes a probability-weighted average. Given a token candidate $w$ with probability $p(w)$, its embedding $\textbf{e}_w$ is computed as:
    %Our contribution towards the reasoning model is a construction of soft embeddings to enforce the model to reason probabilistically. We start by taking probabilistic inputs, tokenize them (convert them to units of text), and compute weighted representations that capture uncertainty in transcription. For each token candidate, it tokenizes the string into potentially multiple subword tokens, retrieves their embeddings, and computes a probability-weighted average. Given a token candidate $w$ with probability $p(w)$, its embedding is computed as:  
    
\begin{equation}
\mathbf{e}_w = p(w) \cdot \frac{1}{|T(w)|} \sum_{t \in T(w)} \mathbf{e}_t
\end{equation}

%\begin{equation}
%\mathbf{e}_w = \mathcal{p}(w) \cdot \frac{1}{|T(w)|} \sum_{t \in T(w)} \mathbf{e}_t
%\end{equation}

where $T(w)$ represents the set of subword tokens obtained from tokenizing $w$, and $\mathbf{e}_t$ is the embedding of a subword token $t$. These embeddings are summed across all candidate words, forming a probabilistic representation. 
The final soft embeddings are stacked into a tensor of shape $[1, N, d]$, where $N$ is the number of soft tokens, and $d$ is the embedding dimension. This formulates our prompt for the language model.
Finally, we concatenate these embeddings with the system prompt embeddings to construct the final input for the model.

%This approach enables us to reason probabilistically. 
%We concatenated embeddings with system prompt embeddings to construct the final input for the model.

%\subsubsection{Reasoning Strategies and Model Optimization}  
%\label{sec:reasoning_strategies}  

%We initiate model inherent reasoning strategies by tuning the model's role description. 

%Instead of directly predicting the final Skill Command, we leverage common-sense reasoning capabilities of the model. Also, we explain the model to expect class objects, where however, the returned object needs to be the real instance. Lastly, we enforce a structured output, ensuring the final word sequence adheres to the predefined Skill Command format. For exact prompt passed to the model, see project website\textsuperscript{\ref{projectwebsiteref}}.

\subsubsection{Scene Embeddings}  
In this work, we consider a fixed scene representation $\mathcal{O}$, consisting of objects and their properties, which is also provided to the language model (LM) as part of its system prompt (see Fig.~\ref{fig:reasoning_prompt}). The scene representation can be extended to handle probabilistic inputs (e.g., uncertainty in object detection and properties from a vision model) or to directly incorporate a scene graph as a structured scene description.
The scene data serve two primary functions: 1) Grounding pointing gestures: Object information helps disambiguate gesture-based selections during preprocessing. 2) Enhancing contextual awareness in the reasoning model: The model receives object properties and attributes (e.g., a small blue cup), ensuring accurate reasoning. When properties are unspecified, the model relies on commonsense reasoning (e.g., assuming a bowl can be used as a container or that a spoon can fit inside a cup).

%We don't consider ambiguous scene representation data in this work, although is could be added also as probabilistic representation. We define constant scene $\mathcal{O}$ and its properties that are given to the LM model as a role description. The scene object data are used to ground pointing gesture object selection. Similarly, it is passed to the system prompt for the Reasoning model to be aware what objects are in the scene and its properties (e.g., a small blue cup).

\section{Preprocessing Gesture and Language}
\label{sec:modality_preprocessing}

This section outlines the preprocessing of hand gestures (Sec.~\ref{sec:hand_gestures_processing}) and voice commands (Sec.~\ref{sec:natural_language_processing}) within the real world experiment (Sec.~\ref{sec:experimental_setup}), transforming them from word-level interpretations acquired from microphone or hand sensor into a probabilistic format ($S_V$ and $S_G$).
Both gestures and language can specify actions. Gestures directly select real-world object instances, whereas voice commands refer to object classes with specified parameters.
%Both modalities contribute equally to the conveyed message, meaning gestures can specify the intended action just as language does. However, a key distinction exists: gestures directly select real-world object instances, whereas voice commands refer to object classes with specified parameters.

\subsection{Hand Gestures}
\label{sec:hand_gestures_processing}

%    \item Gesture Recognition: The hand gesture input is processed using a detection model that extracts relevant features, such as hand pose and motion trajectory. The underlying gesture detection architecture is detailed in Sec.~\ref{sec:hand_gestures_processing}.  

In the real-world experiment, a Leap Motion sensor mounted at the table’s corner captures the hand’s bone structure in real-time (see Fig.~\ref{fig:intro}, bottom left). The Gesture Toolbox \cite{gesturetoolbox} processes this data to recognize individual gestures.
A gesture sentence is recorded while the user’s hand is over the sensor, accumulating gestures throughout an episode and sending them once the episode ends (i.e., when the hand is no longer visible). Each gesture either selects objects (e.g., pointing) or defines actions (e.g., a "fist" for "pick"), based on a predefined mapping.
Both static and dynamic gestures are recognized via cumulative evidence and mapped to target actions $a$ (Sec.~\ref{sec:actions_objects}). Pointing gestures identify objects, assigning probabilities based on their distance from the pointing line \cite{vanc2023communicating}. Each recorded gesture is represented as a probability vector, categorized accordingly (e.g., pointing gestures specify target object $to$). See \cite{gesturetoolbox, vanc2023communicating} for details.

Since gestures can be ambiguous, we assume a one-to-one mapping between each gesture and its Skill Command component (e.g., a "fist" always means "grab"). Users are trained on this mapping, while more generalized mappings were explored in \cite{Vanc2023}.

\subsection{Natural Language}
\label{sec:natural_language_processing}

%\item Speech Recognition: Voice commands are transcribed probabilistically using the \textit{Whisper model} (Sec.~\ref{sec:natural_language_processing}), which produces multiple candidate interpretations with confidence scores.

In the real setup, spoken instructions are converted to text, parsed, filtered, and tokenized. The processed sentence is then matched to predefined language templates (see project website\textsuperscript{\ref{projectwebsiteref}} for the code). We utilized the offline version of OpenAI Whisper model \cite{radford2022whisper}.
Our extended Whisper model inference is transcribing not only single text output but also generating probabilistic representations for each word. It processes the audio through the Whisper model to obtain timestamps, tokens, and corresponding scores, then computes softmax probabilities for alternative tokens. The function extracts the top-$k$ alternatives for each token, applies filtering based on probability thresholds (we use setting $p>0.08$), checks against an English vocabulary, and validates token consistency, ensuring that only credible alternatives remain. Note that while in the simulated dataset, we have fixed the self-defined vocabulary of words, here we work with the whole English vocabulary (NLTK Words corpus~\cite{bird2009natural}). Additionally, it merges subword fragments into complete words to better reflect the intended transcription. These steps provide a richer, uncertainty-aware representation of the transcription compared to the standard, deterministic output of the base Whisper model.

\section{Experimental setup}
\label{sec:experimental_setup}

Our experiments focus on tabletop manipulation tasks in both simulated and real-world settings. For the simulated experiments, we developed a generator of gesture and language commands that allows for model comparison and testing under increased noise and data misalignment (see Sec.~\ref{sec:experiment_noise_align}).
In the real-world experiments, we use the Franka Emika Panda robotic manipulator, equipped with an Intel RealSense D455 RGB-D camera for object perception in a real-world environment (see setup in Fig.~\ref{fig:intro}). Gestures are tracked using a Leap Motion sensor, mounted at the corner of the workspace, and processed via the Gesture Toolbox (see Sec.~\ref{sec:hand_gestures_processing} for details). Voice commands are captured through a microphone and processed using the Whisper model (see Sec.~\ref{sec:natural_language_processing}).

The simulation and real-world setups differ only in how gesture ($S_G$) and language inputs ($S_V$) are generated. The merging algorithm, embedding processing, language model interface, and evaluation of the outputted Skill Command remain unchanged (see Fig.~\ref{fig:system}).

%Our experiments focus on tabletop manipulation tasks using the Franka Emika Panda robotic manipulator. The robot is equipped with an Intel Realsense D455 RGBD camera for object perception operating in real-world environment (see Fig.~\ref{fig:intro}). 

%This paper includes simulated and real experiment. For the artificial one, we created the dataset generator on which we may compare models and test them on increased noise and data disalignment, see Sec.~\ref{sec:experiment_noise_align}.
%For real-world experiment, gestures are tracked using a Leap Motion sensor, mounted at the corner of the workspace. Gesture recognition is detailed in Sec.~\ref{sec:hand_gestures_processing}.

\subsection{Model Benchmarking and Comparisons}
\label{sec:model_comparison}

In our experiments (see Sec.~\ref{sec:exp_results}), we compare transformer-based language models \cite{huggingfacetransformers} fine-tuned for instructions that employ Chain-of-Thought (CoT)~\cite{wei2022chain} to enhance their reasoning capabilities. The models are sourced from the Hugging Face Forum\footnote{Open LLM Leaderboard; Hugging Face Forum: \url{https://huggingface.co/spaces/open-llm-leaderboard/open_llm_leaderboard}, accessed on Feb. 25, 2025.}.
We select models with the highest scores on \textit{IFEval} \cite{zhou2023instructionfollowingevaluationlargelanguage}, prioritizing instruction following and precise formatting. Additionally, to ensure strong language understanding and common-sense reasoning, we consider models with high scores on the BBH dataset \cite{suzgun2022challenging}.
Based on these criteria, we choose three transformer-based models: EXAONE 3.5 2.4B Instruct \cite{exaone_model}, SmolTulu 1.7B Instruct \cite{smoltulu_model}, and Granite 3.1 2B Instruct \cite{granite_model}.
    
%In our experiments (see Sec.~\ref{sec:exp_results}), we compare the transformer-based \cite{huggingfacetransformers} language models finetuned for instructions. These models employ also Chain-of-thoughts \cite{wei2022chain} to improve their reasoning capabilities.  models available at Hugging Face forum\footnote{Open LLM Leaderboard; Hugging Face Forum: \url{https://huggingface.co/spaces/open-llm-leaderboard/open_llm_leaderboard}, accessed on Feb. 25, 2025.}. We choose models with highest score on \textit{IFEval}  \cite{zhou2023instructionfollowingevaluationlargelanguage} to follow instructions and precise formating, also we care about the language understanding and common-sense, therefore we choose based on high score on \textit{BBH} dataset \cite{suzgun2022challenging}. We chose three transformer-based models: \textit{EXAONE 3.5 2.4B Instruct} \cite{exaone_model}, \textit{SmolTulu 1.7b Instruct} \cite{smoltulu_model}, and \textit{Granite 3.1 2B Instruct} \cite{granite_model}.

To evaluate model performance, we compare them against a baseline method. The $Argmax$ baseline follows a greedy decoding strategy, selecting the most probable token for each word, similar to concepts in \cite{jurafskyspeech}. It then constructs the skill command by identifying individual parameters based on their first appearance in the sentence.
%To evaluate their performance, we compared the models with a baseline method. This $Argmax$ baseline directly selects the most probable token for each input, following a greedy strategy for valid token selection, similar to concepts in \cite{jurafskyspeech}.

\subsection{Language model parameters}

The LLM causal model includes a set of tunable parameters, configured as follows: Temperature ($\tau=0$): This keeps the model focused on the most likely outputs, ensuring high precision for structured tasks. Top-p ($top\_p$=1): This controls nucleus sampling, allowing a balance between creativity and precision while maintaining a reasonable token selection. Repetition penalty ($1.1$): This discourages redundant outputs, preventing the model from repeating the same object or action multiple times.
For more details on parameter roles, refer to our code implementation at the project website\textsuperscript{\ref{projectwebsiteref}}.

\subsection{Actions and object set}
\label{sec:actions_objects}
\subsubsection{Objects}  
Object categories for the \textbf{real-world experiment}: \textit{cleaner, bowl, cup, drawer, tomatoes}.  
Object categories for the \textbf{simulated experiment}: \textit{cup, cube, plate, table, can, box, fork, marker, note, storage, blade, rack, ledge, stand, platform}.  
{Object properties}, which help identify objects, include:  
 Size: \textit{small, medium, large},  
Color: \textit{red, green, blue}, and State: \textit{open, closed, half-full}

\subsubsection{Actions}  
Our action space consists of 12 actions, each requiring a different number of parameters:  
{Zero-object actions}: \textit{stop, release, home}.  
{Single-object actions}: \textit{pick, push, pass, place, point, open, close}.  
{Double-object actions}: \textit{pour, put}. Available {prepositions} for these actions include \textit{into} and \textit{onto}, which modify the target of the \textit{pour} and \textit{put} actions. {Adjective action properties} include: \textit{quickly, slowly, carefully, forcefully}, etc.

%\subsubsection{Actions} Our Action space is composed of 12 actions with different amount of required parameters: \textbf{Zero-object actions}: \textit{stop, release, home}. \textbf{Single-object actions}: \textit{pick, push, pass, place, point, open, close}. \textbf{Double-object actions}: \textit{pour, put}.
%Available prepositions for these actions include: into and onto. These prepositions  change the target of actions pour and put. \textbf{Adjective action properties} include: \textit{quickly, slowly, carefully, forcefully, etc.}

%\subsubsection{Objects} Object categories for \textbf{real-world experiment}: \textit{cube, bowl, cup, drawer, bottle}. For \textbf{simulated experiment}: \textit{cup, cube, plate, table, can, box, fork, marker, note, storage, blade, rack, ledge, stand, platform}.
%\textbf{Object properties} that help us identify the object include: {sizes: \textit{small, medium, large}, color: \textit{red, green, blue}, state: \textit{open, closed, half-full}}.

\subsubsection{Robotic skills for individual actions}  
\label{sec:kinesthetic_demos}  

Each action in the real-world experiment is mapped to a corresponding robotic skill, represented as trajectories learned from kinesthetic demonstrations. We build on a learning-from-demonstration (LfD) framework developed by TU Delft~\cite{franka_lfd}, which we ported to ROS2~\cite{franka_lfdROS2}.
For actions involving object manipulation, the target object must be localized before execution. Localization is performed using SIFT-based feature matching, where the detected object is compared against a stored template. Once identified, the system aligns the trajectory with the object's real-time position, ensuring robustness in dynamic environments.
Skill execution is triggered and parametrized by the \textit{Skill Command} (Sec.~\ref{sec:skill_command}) generated by the language model (LM).

%In this work, a skill is defined as a trajectory learned from one or two consecutive kinesthetic demonstrations. Each demonstration is recorded using kinesthetic teaching, where a human physically guides the robot to perform the desired motion. The resulting trajectory is then stored and later retrieved for execution.

%If the skill involves object manipulation, the target object must be localized before execution. Object localization is achieved through SIFT-based feature matching, where the detected object is compared against a stored template. Once the object is identified, the system aligns the trajectory with the object's real-time position, ensuring robustness in dynamic environments. Skill execution is triggered by providing the corresponding Skill Command (Sec.~\ref{sec:skill_command}). We leverage a learning-from-demonstration (LfD) framework, building on the existing implementation developed by TU Delft \cite{franka_learning_from_demonstration}. 

\subsection{Multimodal Artificial Dataset}
\label{sec:artificial_dataset}

We generate an artificial dataset for evaluating the ability to merge the input data under different types of noise (Sec.~\ref{sec:experiment_noise_align}). This dataset synthesizes scene descriptions ($O$), spoken commands with linguistic noise ($S_V$), and gesture input ($S_G$) with temporal misalignment. For each data sample, we randomly select a scene, action, and its parameters, then generate a probabilistic representation of linguistic and gesture commands, following the structure described in Sec.~\ref{sec:probabilistic_modality_representation}. The generator simulates real-world uncertainties, including phonetic errors, filler words, and gesture-to-speech misalignment. Noise parameters include: {phonetic confusion probability $\mathcal{N}_{phon}$, filler words probability $P_{filler}$, alignment noise factor $\mathcal{N}_{align}$, and sentence truncation probability $P_{incomplet}$}. 
 See the project website\textsuperscript{\ref{projectwebsiteref}} for the dataset generation code.%, thereby improving robustness in language-gesture grounding tasks.  

\subsubsection{Scene Object Representation}

Each generated scene consists of multiple unique objects, represented as: $\mathcal{O} = \{o_1, o_2, ..., o_n\}, \quad o_i = \text{{(id, type, properties)}}$, where each object $o_i$ has a \textit{unique identifier}, an \textit{object type}, and \textit{semantic properties} (i.e., size, color, state).  

\subsubsection{Linguistic command generation}

To model \textit{speech errors} that commonly occur in spoken commands, we introduce different types of noise. To model phonetic noise we compute phonetic similarity-based confusion between different words using \textit{FuzzyWuzzy} \cite{fuzzywuzzy} string matching:  

$$
\text{similarity}(w_1, w_2) = \frac{|w_1 \cap w_2|}{|w_1 \cup w_2|} \times 100,
$$
where $w_1$ and $w_2$ are two words. If the similarity score between a word and a confusable counterpart (from NLTK Words corpus~\cite{bird2009natural}) exceeds a predefined threshold ($T_{\text{phonetic}}$), it is randomly substituted with probability ($\mathcal{N}_{phon}$). Additionally, \textit{filler words} (e.g., 'ah', 'like', 'well', etc.) are inserted with a certain probability ($\mathcal{P}_{filler}$), modeling disfluent speech patterns.  Finally, the sentence is truncated with probability $P_{incomplet}$ to simulate missing words.

\subsubsection{Gesture Input Modeling}  

Gestures are generated using a probabilistic model that considers scene context and \textit{temporal misalignment}: 1) The correct object receives a probability value uniformly as $U(0.6, 0.95)$. 2) \textit{Similar-looking objects} (same type) receive distributed probability uniformly as $U(0.2, 0.8)$, modeling possible misinterpretation between objects. 3) A \textit{Gaussian noise model perturbs timestamps} to simulate gesture-to-speech misalignment, with a \textit{shift factor} proportional to the alignment noise level ($\mathcal{N}_{align}$). 
Formally, given a speech timestamp $t_s$, the corresponding gesture timestamp $t_g$ is perturbed as:  

\begin{equation}    
t_g = t_s + \epsilon, \quad \epsilon \sim U(0, 2 \times \mathcal{N}_{align})
\end{equation}

where $U(a, b)$ denotes a uniform distribution.  

\section{Experiment results}
\label{sec:exp_results}

First, we evaluate the models on the simulated dataset from Sec.~\ref{sec:artificial_dataset}, analyzing the impact of individual noise types on their performance (Sec.~\ref{sec:experiment_noise_align}). Second, we conduct a real-world experiment across five different scenarios (Sec.~\ref{sec:exp_results}).
%We vary the levels of individual noise types and assess the performance of the three compared models (see Sec.\ref{sec:model_comparison}) under phonetic noise ($\mathcal{N}_{phon}$) combined with filler words and missing words ($\mathcal{P}_{filler}$ and $P_{incomplet}$). Additionally, we evaluate their robustness to temporal alignment noise ($\mathcal{N}_{align}$).

%Second experiment is a real experiment and check the merging capabilities to prove its usability with comparison to baselines.

%First, we evaluate the models on the simulated dataset from Sec.\ref{sec:artificial_dataset}. We vary the levels of individual noise types and assess the performance of the three compared models (see Sec.\ref{sec:model_comparison}) under phonetic noise ($\mathcal{N}{phon}$) combined with filler and missing words ($\mathcal{P}_{filler}$ and $P_{incomplet}$) and evaluate their robustness to temporal alignment noise ($\mathcal{N}{align}$).

\subsection{Noise Experiment}
\label{sec:experiment_noise_align}

\begin{figure}[h]
    \centering
    \begin{subfigure}[b]{0.28\textwidth}
        \centering
        \includegraphics[width=\textwidth]{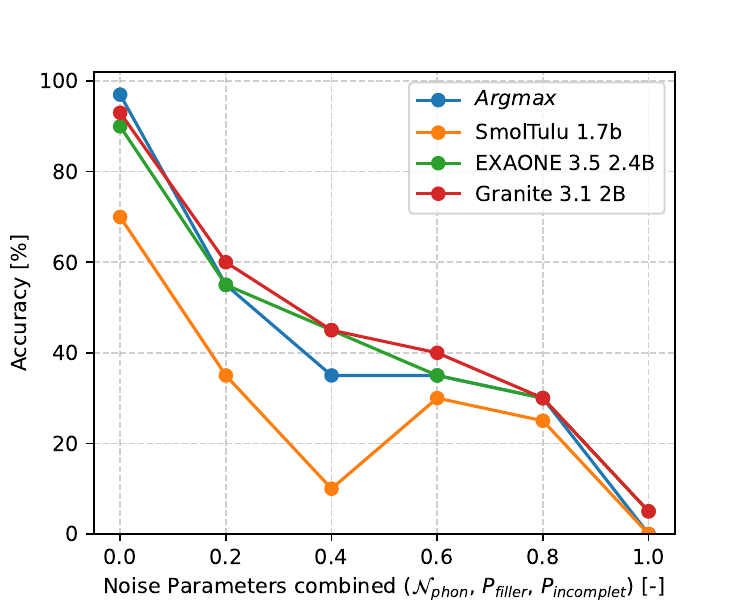}
        \caption{}
        \label{fig:noise2}
    \end{subfigure}
    \hfill
    \begin{subfigure}[b]{0.17\textwidth}
        \centering
        \includegraphics[width=\textwidth]{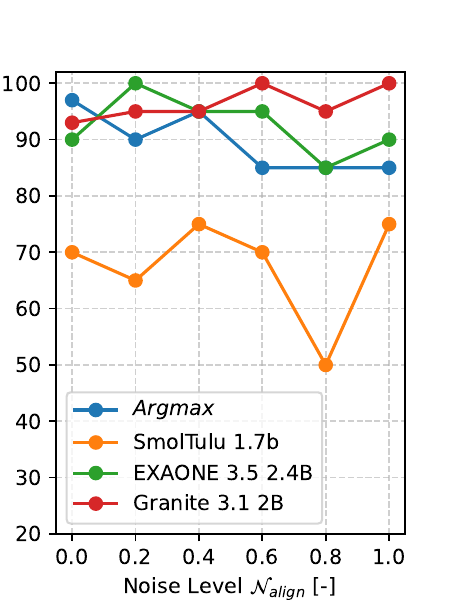}
        \caption{}
        \label{fig:noise1}
    \end{subfigure}
    \caption{(Left) Influence of combined noises on model's performance ($\mathcal{P}_{filler}=\mathcal{P}_{filler}=\mathcal{N}_{phon}$). (Right) The Alignment noise increase alone doesn't affect the accuracy. Each point is average of 20 samples.
    }
    \label{fig:noise}
\end{figure}

First, we evaluate the models on the simulated dataset from Sec.~\ref{sec:artificial_dataset}. We assess the performance of four models—three state-of-the-art models and the Argmax baseline (see Sec.~\ref{sec:model_comparison})—under phonetic noise ($\mathcal{N}_{phon}$) combined with filler word ($P_{filler}$) and missing word probabilities ($P_{incomplet}$). In this setting, all noise levels are set to the same value.
As shown in Fig.\ref{fig:noise2}, all models are affected by the combined noise. As expected, at zero noise, the Argmax model outperforms TransforMerger. However, as noise increases, TransforMerger with the Granite~\cite{granite_model} reasoning engine surpasses Argmax, reaching ~97\% accuracy at zero noise and ~40\% at noise level 0.6. The SmolTulu model~\cite{smoltulu_model} performed the worst, even underperforming Argmax, though it still achieved reasonable results (~70\% accuracy at zero noise, 30\% at noise level 0.6).  

Next, we evaluate model robustness to temporal alignment noise ($\mathcal{N}_{align}$). As shown in Fig.~\ref{fig:noise1}, misalignment had impacted the Argmax model but had little to no effect on TransforMerger’s performance. Both the Granite and EXAONE reasoning engines outperformed Argmax, achieving nearly 100\% accuracy. Note that the misalignment was limited to a single position; greater discrepancies could lead to larger performance drops for Argmax, as observed in real-world experiments.%while each object and action was described by multiple words (e.g., object properties, filler words). As a result, the object order remained typically unchanged, enabling proper word linking for the Argmax model, which only failed when a word was missing from one modality or when the word order for objects was altered. Meanwhile, the language models inferred alignment directly from raw input and performed exceptionally well, with Granite model~\cite{granite_model} and EXAONE model~\cite{exaone_model} achieving the best results close to 100\% accuracy.

We increase the phonetic confusion probability $\mathcal{N}_{phon}$, filler words probability $P_{filler}$, and sentence truncation probability $P_{incomplet}$, then we compared different models with baselines, see Fig.~\ref{fig:noise1}. Secondly, we increase the  alignment noise factor $\mathcal{N}_{align}$, see Fig.~\ref{fig:noise2}.

% \begin{figure}[h]
%     \centering
%     \begin{subfigure}[b]{0.3\textwidth}
%         \centering
%         \includegraphics[width=0.9\textwidth]{experiment_figures/noise_plot1_newdataset.pdf}
%         \caption{}
%         \label{fig:noise2}
%     \end{subfigure}
%     \hfill
%     \begin{subfigure}[b]{0.16\textwidth}
%         \centering
%         \includegraphics[width=\textwidth]{experiment_figures/noise_plot2_newdataset.pdf}
%         \caption{}
%         \label{fig:noise1}
%     \end{subfigure}
%     \caption{(Left) Influence of combined noises on model's performance ($\mathcal{P}_{filler}=0.1$ and $\mathcal{P}_{filler}=0.1$). (Right) The Alignment noise increase alone doesn't affect the accuracy. Each point is average of 20 samples.
%     }
%     \label{fig:noise}
% \end{figure}

\subsection{Real Experiment}
\label{sec:exp_results}
The goal of this experiment is to evaluate the system’s performance under various conditions in real-world interactions (see the attached video for individual usecases). We evaluated the following scenarios:
\begin{enumerate}
\item $T_1$: Evaluating effect of partial or noisy inputs. Command: 'Pick cube' + gestures + scene. 
\item $T_2$: Evaluating ability to resolve ambiguity in language object description. Command: 'Pick the red object' + gestures + scene with two red objects. 
\item $T_3$: Evaluate the role of contextual information and commonsense reasoning in disambiguating objects when an action involves two parameters. Command: "Put cube to box"+ gestures + scene. 
\item $T_4$: Evaluate how the models can compensate for missing information by using multimodal inputs. Command: "Put this to that" + gestures + scene. 
\end{enumerate}
%First, we assess how the system handles partial or noisy inputs, such as voice or gesture errors, similar to the simulated experiment (scenario $T_1$, Command: 'Pick cube' + corresponding gesture). Second, we examine its ability to resolve ambiguity in language object description ($T_2$, Command: 'Pick the red object' + scene with two red objects). Third, we evaluate the role of contextual information and commonsense reasoning in disambiguating objects when an action involves two parameters (Scenario $T_3$, Command: "Put cup to bowl"). Finally, we evaluate how the models can compensate for missing information by leveraging multimodal inputs ($T_4$, Command: "Put THIS to THAT" + two gesture pointings).

\begin{figure}
    \centering
    \includegraphics[width=0.97\linewidth]{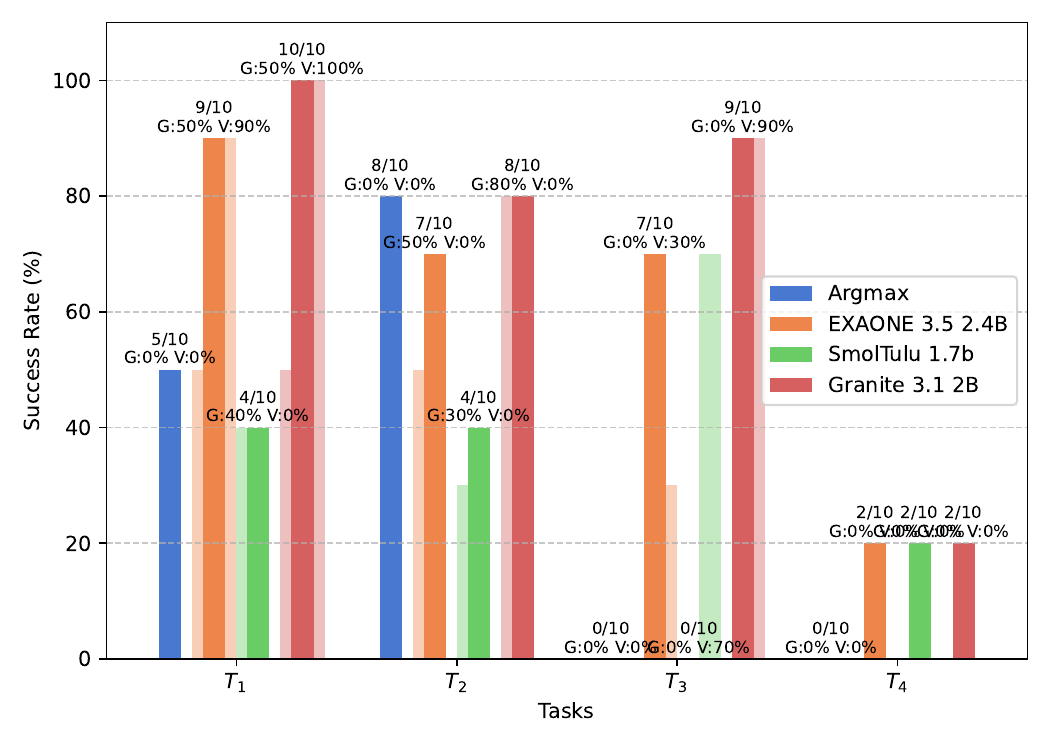}
    \caption{%(Alternative with single modality data, experiment\_figures/plotter.py to plot these) Handcrafted scenarios, 
    Results of the real experiment for tasks $T_{\{1,2,3,4\}}$, each task has 10 executions. Light color shows the result for a single modality input, left-to-bar is the gestures only, right-to-bar is the voice command only.}
    \label{fig:real-world}
\vspace{-0.5cm}
\end{figure}
The results for individual scenarios are shown in Fig.~\ref{fig:real-world}, with each scenario repeated 10 times. Key observations include: 
1) SmolTulu consistently performed the worst (worse than Argmax for $T_1$ and $T_2$), Granite the best, and Exaone a close second. These results are mirroring the simulated experiments. 
2) Argmax performed worse under natural noise ($T_1$) than in simulation, suggesting that simulated noise levels were lower than real-world noise. However, . However, Granite succeeded in all 10 trials, even under significant real-world noise, whereas SmolTulu succeeded in only 40\% of trials. Granite was able to resolve this task using language alone in all cases and using gestures alone in 5 cases. Argmax failed when relying on a single modality. Granite was able to resolve this task also by language only in all the cases and in 5 cases also by gestures only. 
3) $T_2$ is unresolvable using language alone. Both Argmax and Granite performed well with multimodal input. However, thanks to its soft embeddings, Granite achieved the same accuracy using only gesture commands as it did with multimodal input, whereas Argmax failed due to imprecise pointing. 
4) In $T_3$, Argmax completely failed, as it could not infer the intended object based on indirect descriptions. In contrast, Granite performed equally well using voice alone and multimodal input, leveraging contextual knowledge to resolve object ambiguity without requiring gestures. %Despite the lack of explicit scene information, language models' commonsense reasoning compensated effectively, leading to performance similar to $T_2$. 
5) $T_4$ which involved resolving two missing parameters in language commands, was the most challenging scenario. Only two out of ten trials were successful across all models (except Argmax). Additionally, using only voice or only gestures resulted in 0\% accuracy for this scenario for all of the models, highlighting the importance of multimodal merging combined with good reasoning capabilities.% This highlights the usefulness of the proposed approach, which enables individual modalities to complement each other and compensate for noise in individual modalities due to probabilistic merging together with contextual and commonsense information.

\section{Conclusion}

In this paper, we introduced TransforMerger, a novel approach that leverages transformer-based language models tuned for instruction-following and reasoning in multimodal human-robot interaction. We demonstrated how noisy, ambiguous (\textit{Put the red object into green object}), or incomplete language and gesture commands (\textit{Pour this there}) can be probabilistically merged using contextual knowledge to generate parametrized skill commands for robotic execution.

Both simulated and real-world experiments confirmed that these models can effectively process probabilistic multimodal inputs. When provided with contextual information and available actions, the best-performing models often resolved ambiguous or noisy commands, directly generating executable skill commands.
Performance varied significantly among the top three models, with Granite reaching up to 100\% accuracy in low-noise scenarios, consistently outperforming SmolTulu, often by a factor of two. However, even SmolTulu performed surprisingly well in low-noise conditions.

Our real-world experiment demonstrated the full pipeline—from language and gesture human input to task execution—highlighting the effectiveness of probabilistic merging and commonsense reasoning in enabling robust multimodal interaction. While the models still struggle in certain cases, failing to fully adhere to specific reasoning rules or correctly interpret probabilistic results, they already achieve promising performance, surpassing deterministic approached especially in more ambiguous scenarios. This underscores a promising path toward more natural multimodal communication across diverse contexts, where individual modalities complement each other, compensating for misalignment and noise through probabilistic merging, contextual grounding, and commonsense reasoning.

%This study underscores the usefulness of the proposed approach, where individual modalities complement each other, compensating for noise through probabilistic merging, contextual grounding, and commonsense reasoning.

%The probability distribution with items, where each item is represented as the single valid token of the model's tokenizer.

%Alignment - 
%Alignment noise in the simulated dataset might have been too little. The other noises successfully imitated 

\bibliographystyle{IEEEtran}
\bibliography{main}

% Generated by IEEEtran.bst, version: 1.14 (2015/08/26)
\begin{thebibliography}{10}
\providecommand{\url}[1]{#1}
\csname url@samestyle\endcsname
\providecommand{\newblock}{\relax}
\providecommand{\bibinfo}[2]{#2}
\providecommand{\BIBentrySTDinterwordspacing}{\spaceskip=0pt\relax}
\providecommand{\BIBentryALTinterwordstretchfactor}{4}
\providecommand{\BIBentryALTinterwordspacing}{\spaceskip=\fontdimen2\font plus
\BIBentryALTinterwordstretchfactor\fontdimen3\font minus \fontdimen4\font\relax}
\providecommand{\BIBforeignlanguage}[2]{{%
\expandafter\ifx\csname l@#1\endcsname\relax
\typeout{** WARNING: IEEEtran.bst: No hyphenation pattern has been}%
\typeout{** loaded for the language `#1'. Using the pattern for}%
\typeout{** the default language instead.}%
\else
\language=\csname l@#1\endcsname
\fi
#2}}
\providecommand{\BIBdecl}{\relax}
\BIBdecl

\bibitem{pires2005robot}
J.~N. Pires, ``Robot-by-voice: experiments on commanding an industrial robot using the human voice,'' \emph{Industrial Robot: An International Journal}, vol.~32, no.~6, pp. 505--511, 2005.

\bibitem{Vanc2023}
P.~Vanc, J.~K. Behrens, and K.~Stepanova, ``Context-aware robot control using gesture episodes,'' in \emph{2018 IEEE/RSJ ICRA}, 2023.

\bibitem{Wang_2024}
C.~Wang, S.~Hasler, D.~Tanneberg \emph{et~al.}, ``Lami: Large language models for multi-modal human-robot interaction,'' in \emph{Extended Abstracts of the CHI Conference on Human Factors in Computing Systems}, ser. CHI ’24.\hskip 1em plus 0.5em minus 0.4em\relax ACM, May 2024, p. 1–10.

\bibitem{huggingfacetransformers}
T.~Wolf, L.~Debut, V.~Sanh \emph{et~al.}, ``Huggingface's transformers: State-of-the-art natural language processing,'' Oct. 2020.

\bibitem{bishop2006pattern}
C.~M. Bishop and N.~M. Nasrabadi, \emph{Pattern recognition and machine learning}.\hskip 1em plus 0.5em minus 0.4em\relax Springer, 2006, vol.~4, no.~4.

\bibitem{rabiner1989tutorial}
L.~R. Rabiner, ``A tutorial on hidden markov models and selected applications in speech recognition,'' \emph{Proceedings of the IEEE}, vol.~77, no.~2, pp. 257--286, 1989.

\bibitem{starner1998visual}
T.~Starner, B.~Schiele, and A.~Pentland, ``Visual contextual awareness in wearable computing,'' in \emph{Digest of Papers. Second International Symposium on Wearable Computers (Cat. No. 98EX215)}.\hskip 1em plus 0.5em minus 0.4em\relax IEEE, 1998, pp. 50--57.

\bibitem{gpttrasnformersbrown2020}
T.~B. Brown, B.~Mann, N.~Ryder \emph{et~al.}, ``Language models are few-shot learners,'' \emph{Advances in neural information processing systems}, vol.~33, pp. 1877--1901, 2020.

\bibitem{wei2022finetunedlanguagemodelszeroshot}
\BIBentryALTinterwordspacing
J.~Wei, M.~Bosma, V.~Y. Zhao, K.~Guu, A.~W. Yu, B.~Lester, N.~Du, A.~M. Dai, and Q.~V. Le, ``Finetuned language models are zero-shot learners,'' 2022. [Online]. Available: \url{https://arxiv.org/abs/2109.01652}
\BIBentrySTDinterwordspacing

\bibitem{dosovitskiy2021imageworth16x16words}
A.~Dosovitskiy, L.~Beyer, A.~Kolesnikov \emph{et~al.}, ``An image is worth 16x16 words: Transformers for image recognition at scale,'' in \emph{ICLR}, 2021.

\bibitem{radford2022robustspeechrecognitionlargescale}
A.~Radford, J.~W. Kim, T.~Xu \emph{et~al.}, ``Robust speech recognition via large-scale weak supervision,'' PMLR, pp. 28\,492--28\,518, 2023.

\bibitem{NEURIPS2022_960a172b}
J.-B. Alayrac, J.~Donahue, P.~Luc \emph{et~al.}, ``Flamingo: a visual language model for few-shot learning,'' \emph{Advances in neural information processing systems}, vol.~35, pp. 23\,716--23\,736, 2022.

\bibitem{wang2024thisthatlanguagegesturecontrolledvideo}
\BIBentryALTinterwordspacing
B.~Wang, N.~Sridhar, C.~Feng \emph{et~al.}, ``{T}his\&{T}hat: Language-gesture controlled video generation for robot planning,'' 2024. [Online]. Available: \url{https://arxiv.org/abs/2407.05530}
\BIBentrySTDinterwordspacing

\bibitem{trick2019multimodaluncertaintyreductionintention}
S.~Trick, D.~Koert, J.~Peters, and C.~Rothkopf, ``Multimodal uncertainty reduction for intention recognition in human-robot interaction,'' in \emph{IEEE/RSJ Int. Conf. on Intelligent Robots and Systems (IROS)}, 2019.

\bibitem{ferrari2024collaborativeconversationsafemultimodal}
D.~Ferrari, A.~Pupa, and C.~Secchi, ``Collaborative conversation in safe multimodal human-robot collaboration,'' in \emph{IEEE Int. Conf. on Intelligent Robots and Systems (IROS)}, 2024.

\bibitem{lai2025nmmhrinaturalmultimodalhumanrobot}
\BIBentryALTinterwordspacing
Y.~Lai, S.~Yuan, Y.~Nassar \emph{et~al.}, ``Nmm-hri: Natural multi-modal human-robot interaction with voice and deictic posture via large language model,'' 2025. [Online]. Available: \url{https://arxiv.org/abs/2501.00785}
\BIBentrySTDinterwordspacing

\bibitem{baltruvsaitis2018multimodal}
T.~Baltru{\v{s}}aitis, C.~Ahuja, and L.-P. Morency, ``Multimodal machine learning: A survey and taxonomy,'' \emph{IEEE transactions on pattern analysis and machine intelligence}, vol.~41, no.~2, pp. 423--443, 2018.

\bibitem{gesturetoolbox}
P.~Vanc, ``{Gesture teleoperation toolbox v.0.1},'' \url{https://github.com/imitrob/teleop gesture toolbox/}, 2022.

\bibitem{vanc2023communicating}
P.~Vanc, J.~K. Behrens, K.~Stepanova, and V.~Hlavac, ``Communicating human intent to a robotic companion by multi-type gesture sentences,'' in \emph{IEEE/RSJ IROS}, 2023, pp. 9839--9845.

\bibitem{radford2022whisper}
A.~Radford, J.~W. Kim, T.~Xu \emph{et~al.}, ``Robust speech recognition via large-scale weak supervision,'' \url{https://github.com/openai/whisper}, 2022, accessed: 2025-02-25.

\bibitem{bird2009natural}
S.~Bird, E.~Loper, and E.~Klein, \emph{Natural Language Processing with Python}.\hskip 1em plus 0.5em minus 0.4em\relax O'Reilly Media Inc., 2009.

\bibitem{wei2022chain}
J.~Wei, X.~Wang, D.~Schuurmans \emph{et~al.}, ``Chain-of-thought prompting elicits reasoning in large language models,'' \emph{Advances in neural information processing systems}, vol.~35, pp. 24\,824--24\,837, 2022.

\bibitem{zhou2023instructionfollowingevaluationlargelanguage}
\BIBentryALTinterwordspacing
J.~Zhou, T.~Lu, S.~Mishra \emph{et~al.}, ``Instruction-following evaluation for large language models,'' 2023. [Online]. Available: \url{https://arxiv.org/abs/2311.07911}
\BIBentrySTDinterwordspacing

\bibitem{suzgun2022challenging}
M.~Suzgun, N.~Scales, N.~Sch{\"a}rli \emph{et~al.}, ``Challenging big-bench tasks and whether chain-of-thought can solve them,'' \emph{ACL}, 2023.

\bibitem{exaone_model}
\BIBentryALTinterwordspacing
L.~A. Research, S.~An, K.~Bae, E.~Choi \emph{et~al.}, ``Exaone 3.5: Series of large language models for real-world use cases,'' 2024. [Online]. Available: \url{https://arxiv.org/abs/2412.04862}
\BIBentrySTDinterwordspacing

\bibitem{smoltulu_model}
\BIBentryALTinterwordspacing
S.~Alrashed, ``Smoltulu: Higher learning rate to batch size ratios can lead to better reasoning in slms,'' 2024. [Online]. Available: \url{https://arxiv.org/abs/2412.08347}
\BIBentrySTDinterwordspacing

\bibitem{granite_model}
\BIBentryALTinterwordspacing
I.~Granite, ``Granite 3.1 language models,'' 2025, accessed: 2025-02-28. [Online]. Available: \url{https://github.com/ibm-granite/granite-3.1-language-models/}
\BIBentrySTDinterwordspacing

\bibitem{jurafskyspeech}
J.~H. Martin and D.~Jurafsky, \emph{Speech and language processing: An introduction to natural language processing, computational linguistics, and speech recognition}.\hskip 1em plus 0.5em minus 0.4em\relax Pearson/Prentice Hall Upper Saddle River, 2009, vol.~23.

\bibitem{franka_lfd}
G.~Francesze, ``Franka learning from demonstrations,'' \url{https://github.com/platonics-delft/franka_learning_from_demonstrations}, 2024, {P}latonics {D}elft, Accessed: Mar. 1, 2025.

\bibitem{franka_lfdROS2}
P.~Vanc, ``Franka learning from demonstrations - {ROS2},'' \url{https://github.com/imitrob/franka_learning_from_demonstrations_ros2}, 2024, accessed: Mar. 1, 2025.

\bibitem{fuzzywuzzy}
I.~SeatGeek, ``Fuzzywuzzy: Fuzzy string matching in python,'' \url{https://github.com/seatgeek/fuzzywuzzy}, 2011, accessed: 2025-02-25.

\end{thebibliography}

%\appendix

\end{document}